\pdfoutput=1
\documentclass{article}
\usepackage{spconf,amsmath,graphicx}
\usepackage{amsmath}  
\usepackage{color}
\usepackage{algorithm}  
\usepackage[noend]{algpseudocode}   
\usepackage{xcolor}
\usepackage{enumerate}
\usepackage{multirow}
\usepackage{float}
\usepackage{booktabs}
\usepackage{xcolor}
\usepackage{graphicx}  
\usepackage{float}  
\usepackage{subfigure}  
\usepackage{amssymb}

\title{A Comprehensive Comparison of Projections in Omnidirectional Super-Resolution}
\name{Huicheng Pi\textsuperscript{1, \textdagger} \quad Senmao Tian\textsuperscript{1,5, \textdagger}\thanks{{\textdagger}  Equal contribution.}\quad Ming Lu\textsuperscript{2} \quad Jiaming Liu\textsuperscript{3,5} \quad Yandong Guo\textsuperscript{4} \quad Shunli Zhang\textsuperscript{1,*}\thanks{{*} This work was supported by the Fundamental Research Funds for the Central Universities (2022JBMC013), the National Natural Science Foundation of China (61976017 and 61601021), and the Beijing Natural Science Foundation (4202056). Shunli Zhang is the corresponding author.}}
\address{
	\textsuperscript{1} School of Software Engineering, Beijing Jiaotong University, 
	\textsuperscript{2} Intel Labs China, \textsuperscript{3} Peking University \\\textsuperscript{4} Beijing University of Posts and Telecommunications, \textsuperscript{5} OPPO Research Institute}

\address{
	\textsuperscript{1} School of Software Engineering, Beijing Jiaotong University, 
	\textsuperscript{2} Intel Labs China, \textsuperscript{3} Peking University \\\textsuperscript{4} Beijing University of Posts and Telecommunications, \textsuperscript{5} OPPO Research Institute}

\begin{document}
%
\maketitle
\begin{abstract}
Super-Resolution (SR) has gained increasing research attention over the past few years. With the development of Deep Neural Networks (DNNs), many super-resolution methods based on DNNs have been proposed. Although most of these methods are aimed at ordinary frames, there are few works on super-resolution of omnidirectional frames. In these works, omnidirectional frames are projected from the 3D sphere to a 2D plane by Equi-Rectangular Projection (ERP). Although ERP has been widely used for projection, it has severe projection distortion near poles. Current DNN-based SR methods use 2D convolution modules, which is more suitable for the regular grid. In this paper, we find that different projection methods have great impact on the performance of DNNs. To study this problem, a comprehensive comparison of projections in omnidirectional super-resolution is conducted. We compare the SR results of different projection methods. Experimental results show that Equi-Angular cube map projection (EAC), which has minimal distortion, achieves the best result in terms of WS-PSNR compared with other projections. Code and data will be released.
\end{abstract}

\begin{figure*}[h]
	\centering
	\includegraphics[width=0.8\textwidth,height=0.4\textwidth]{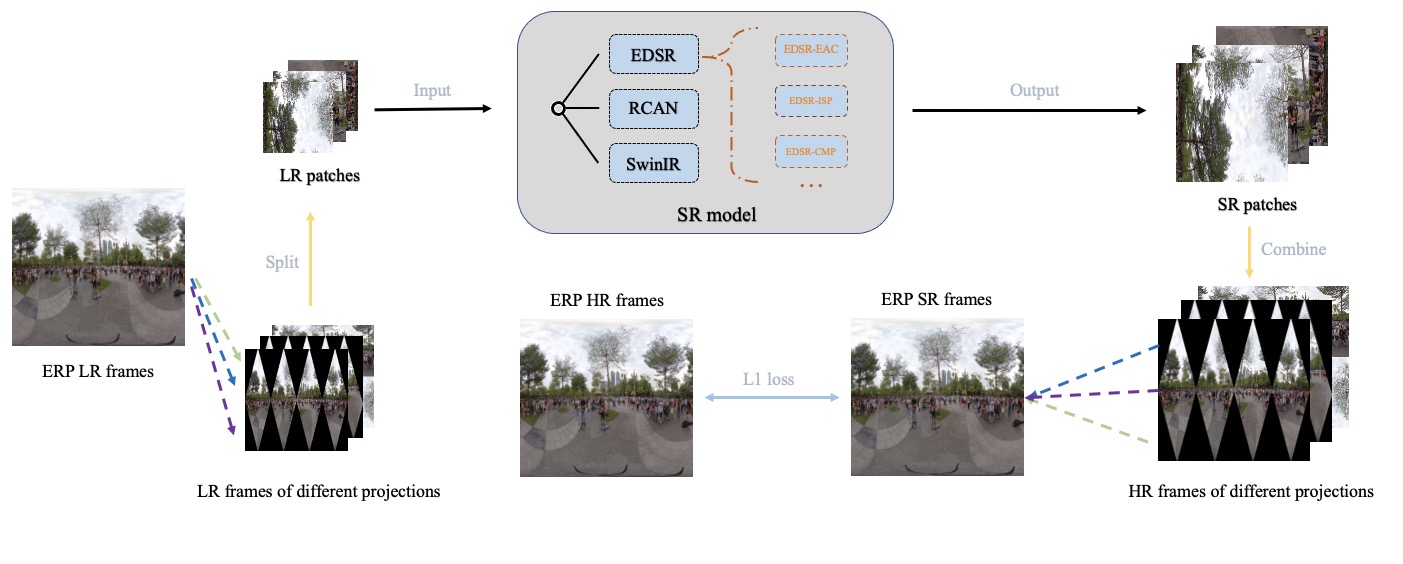}
	\caption{The pipeline of our method.}
	\label{fig:pipeline}
\end{figure*}

\begin{figure*}[h]
	\centering
	\subfigure[ERP]{
		\includegraphics[width=2.2cm]{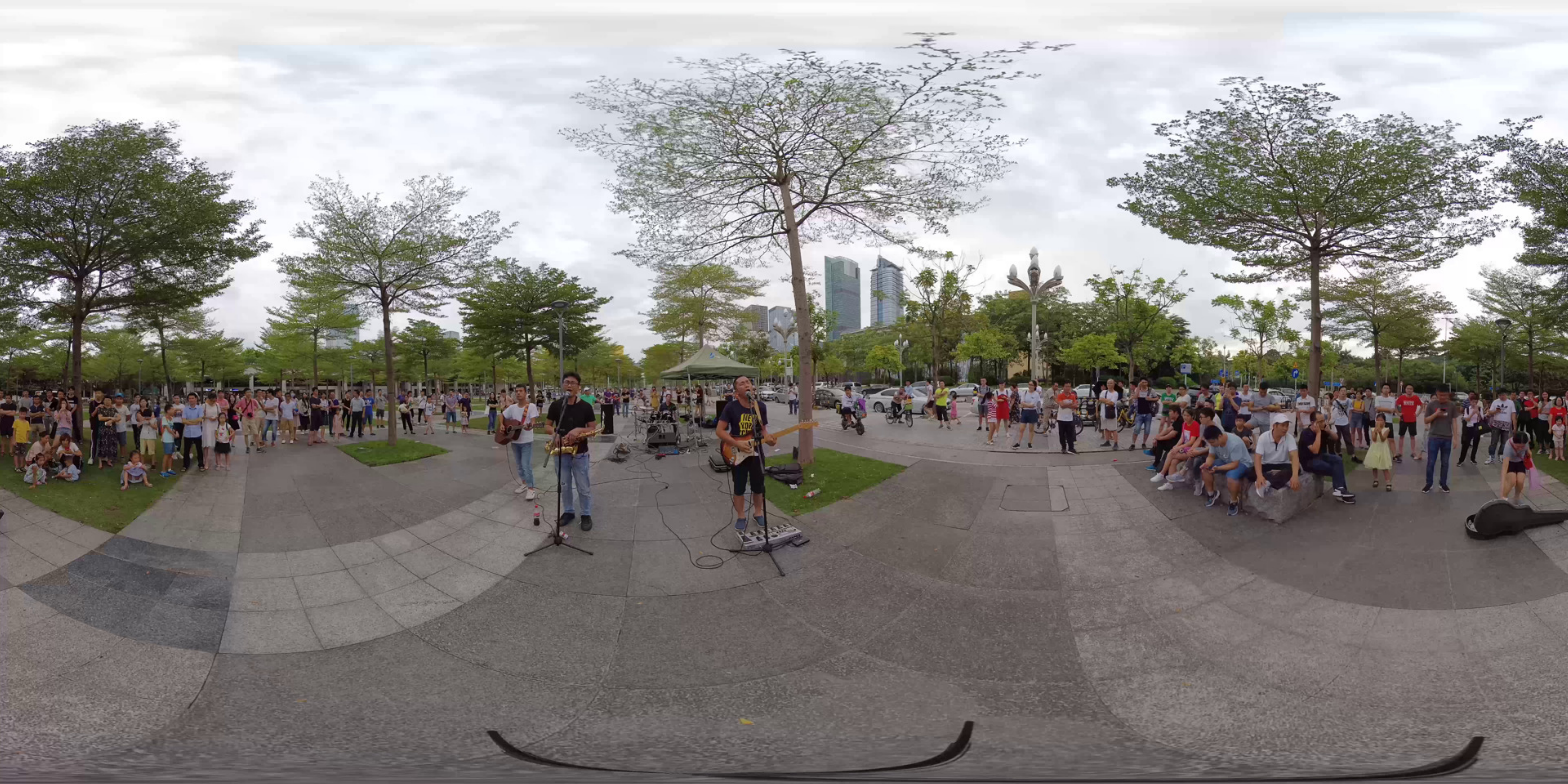}
	}
	\subfigure[EAC]{
		\includegraphics[width=2.2cm]{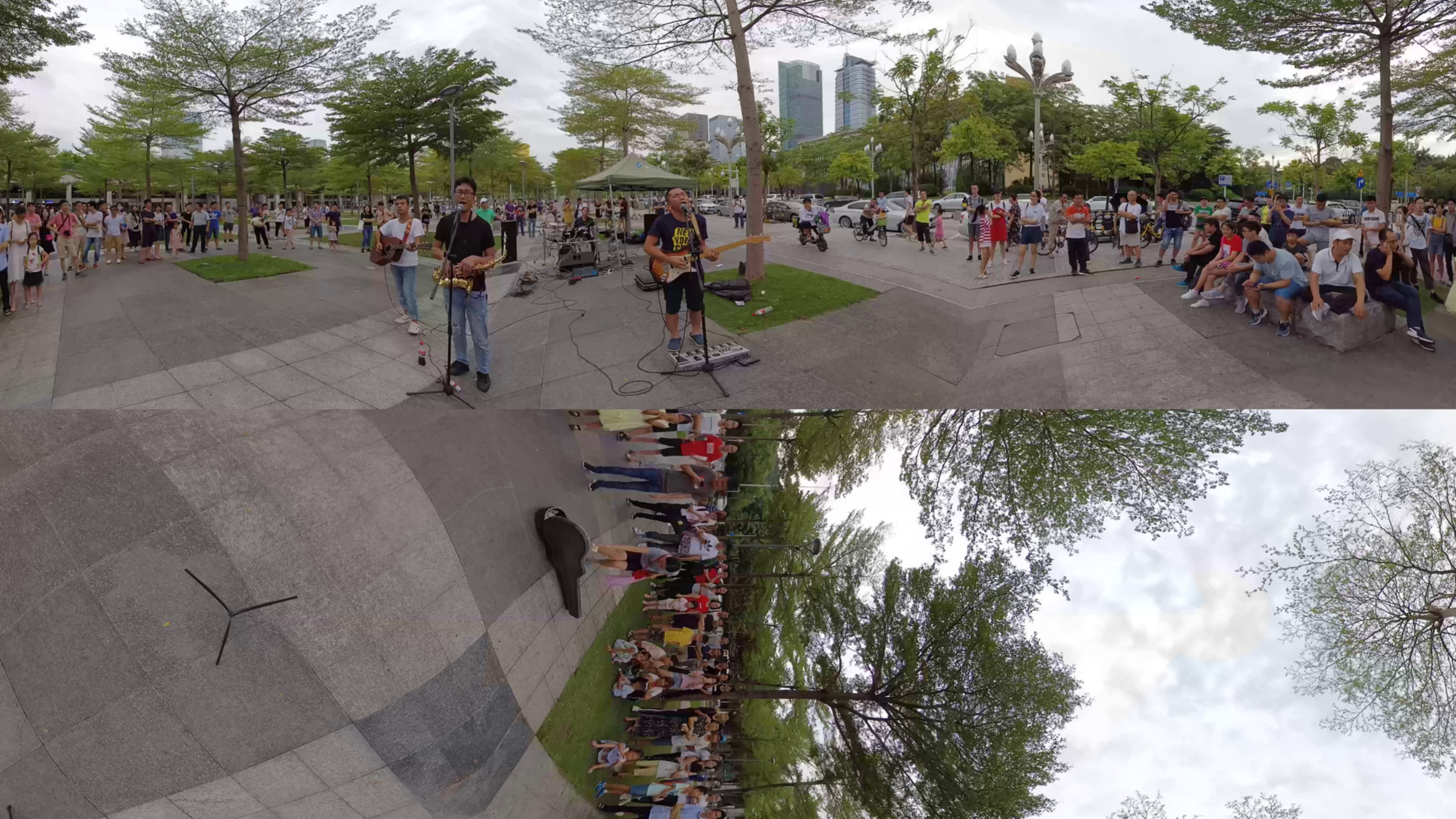}
	}
	\subfigure[CMP]{
		\includegraphics[width=2.2cm]{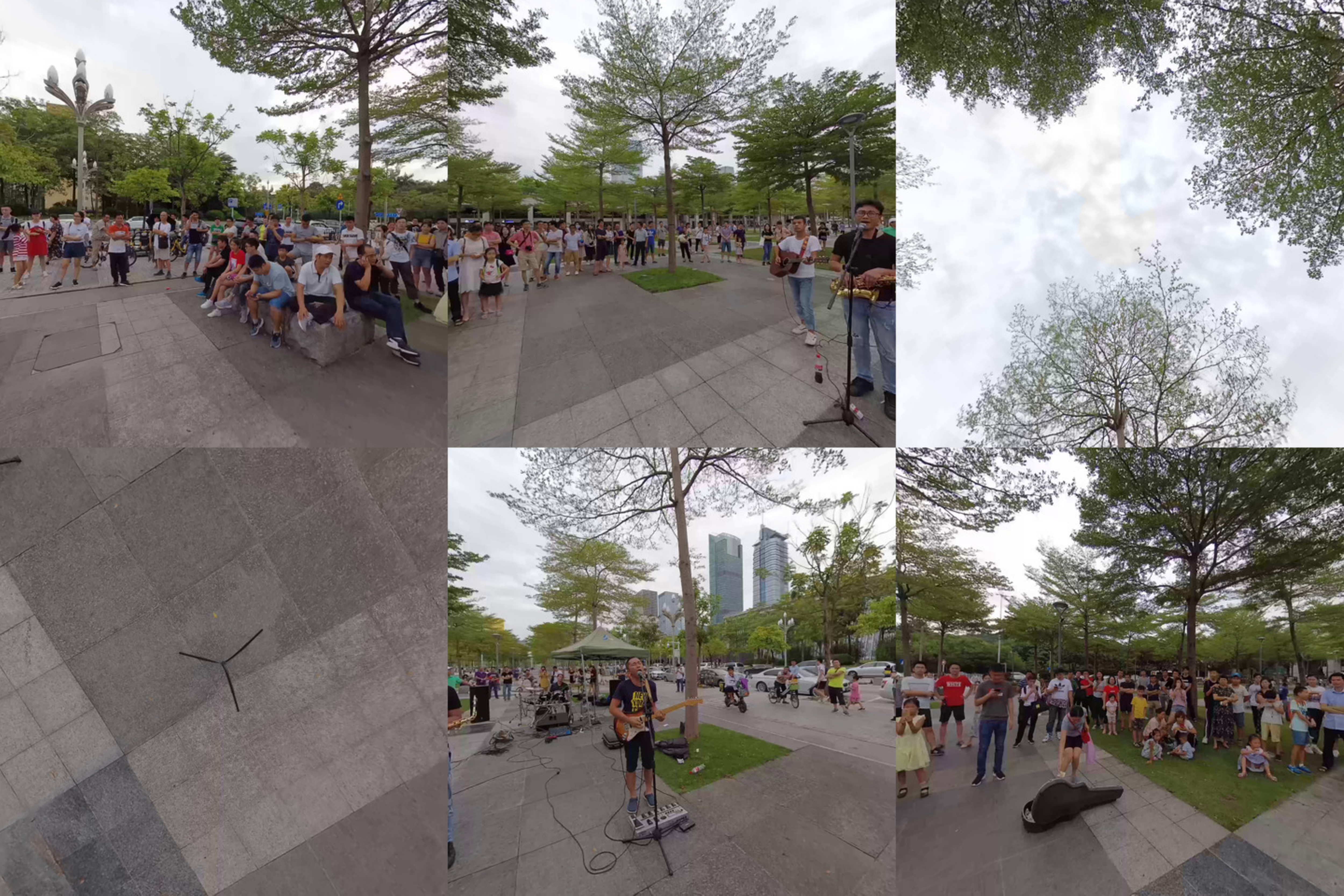}
	}
	\subfigure[SSP]{
		\includegraphics[width=2.2cm]{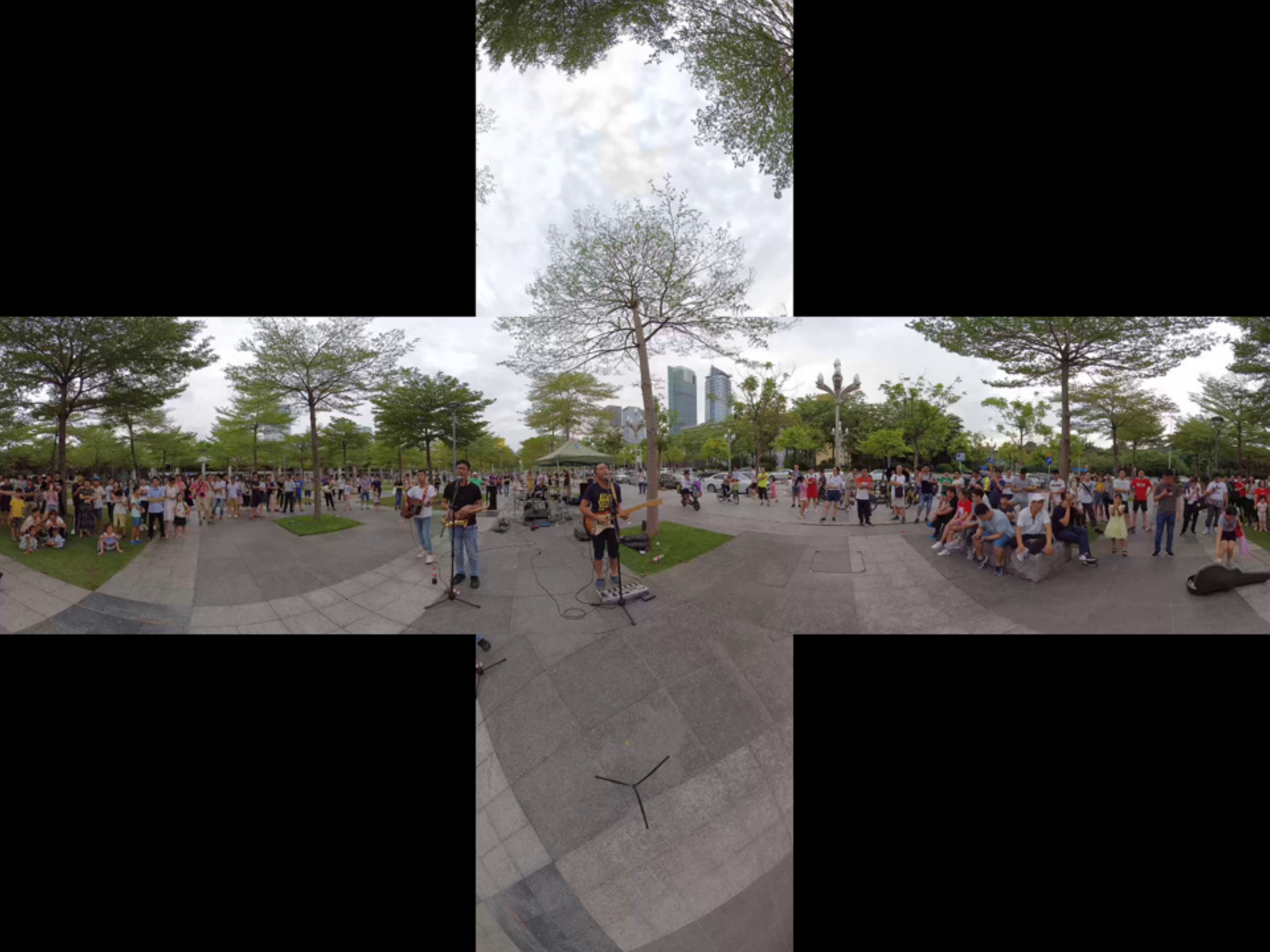}
	}
	\subfigure[ISP]{
		\includegraphics[width=2.2cm]{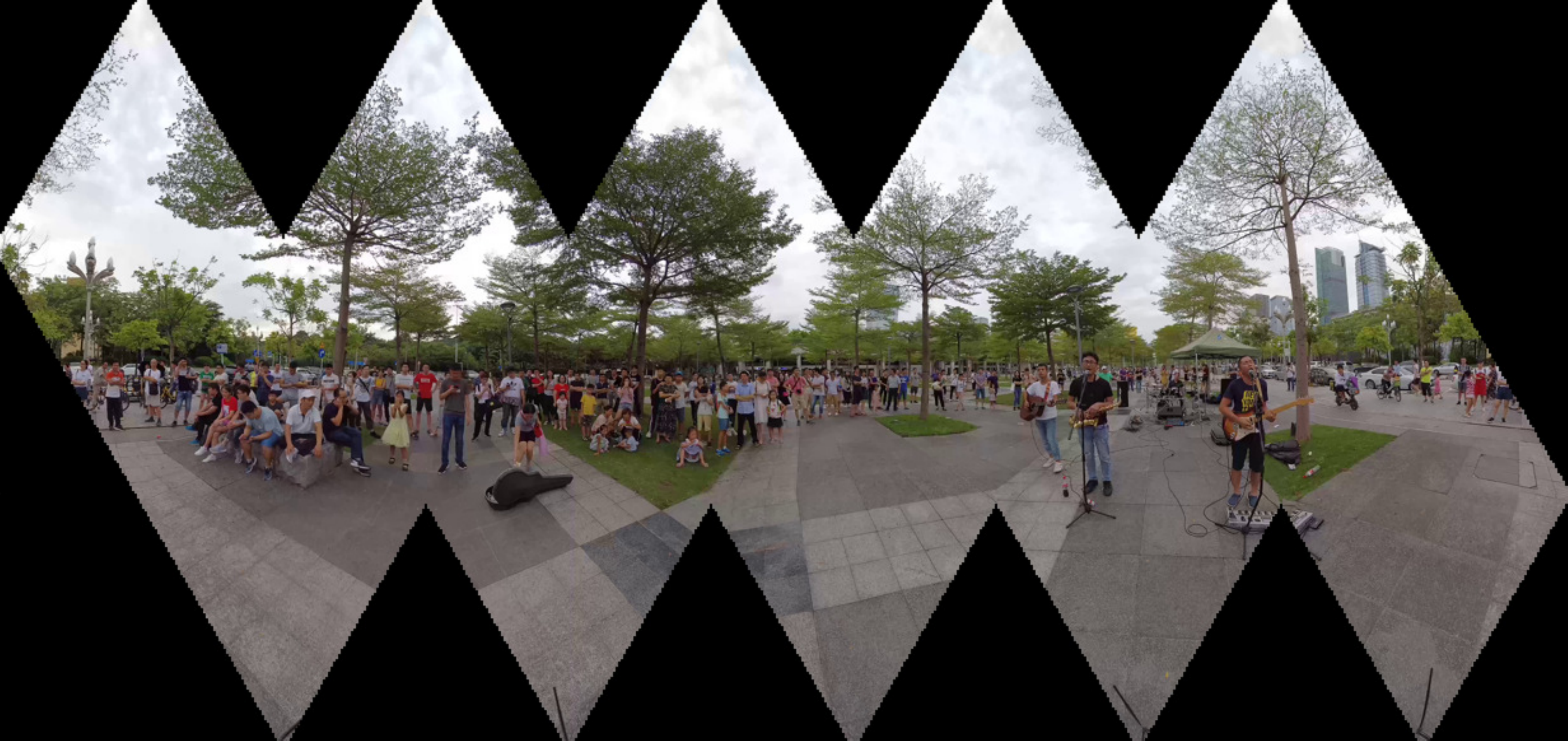}
	}
	\subfigure[TSP]{
		\includegraphics[width=2.2cm]{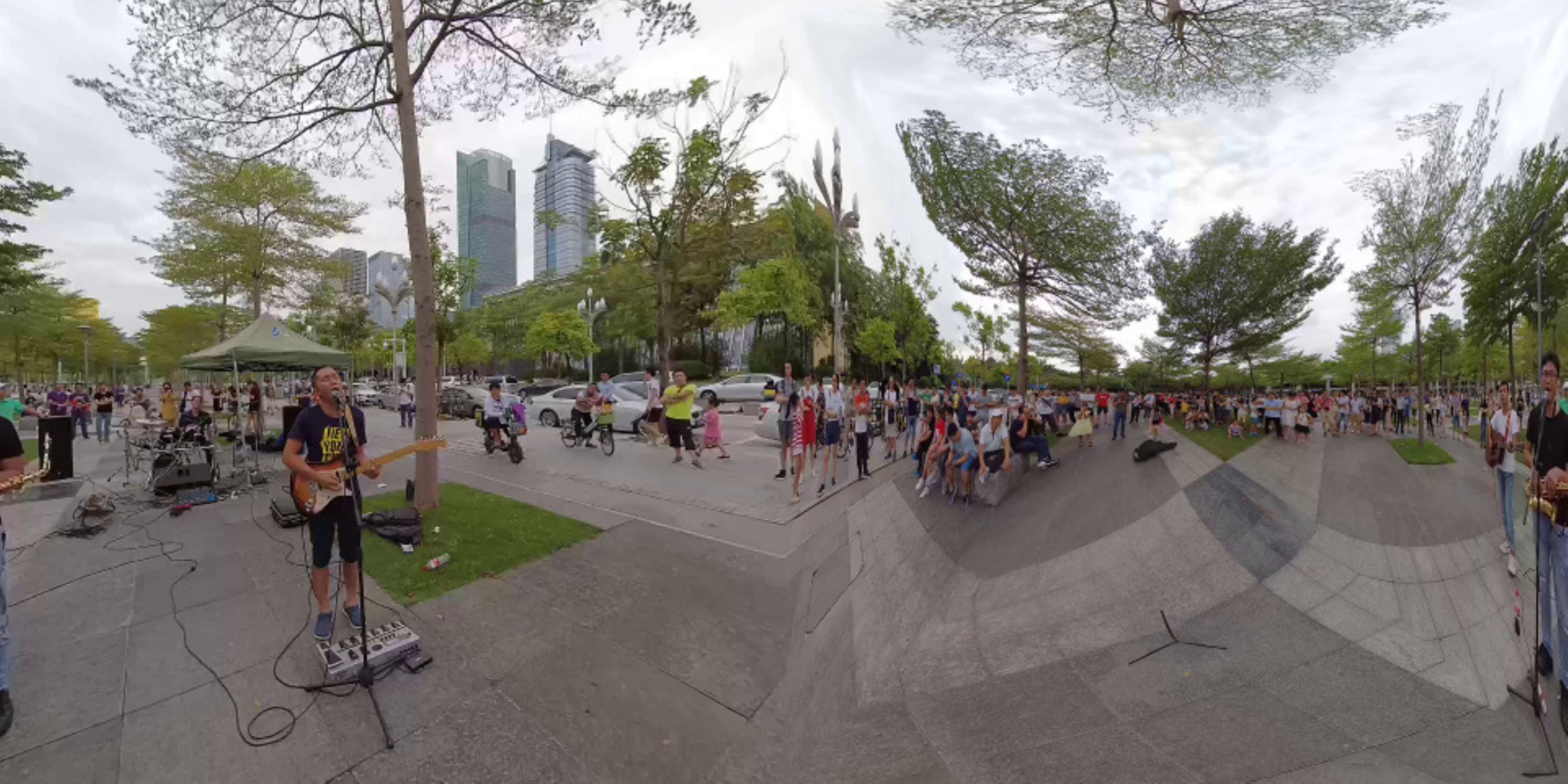}
	}
	\subfigure[OHP]{
		\includegraphics[width=2.2cm]{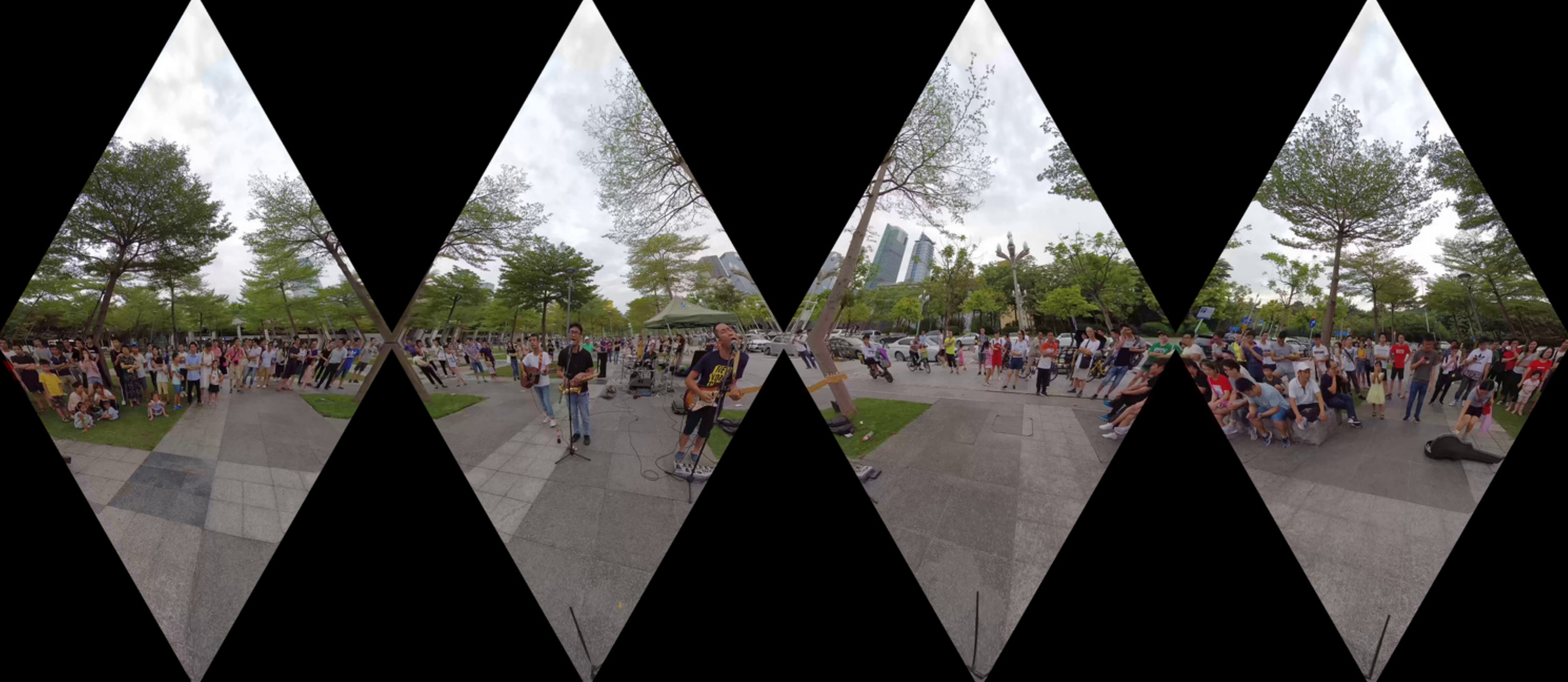}
	}
	\caption{Illustration of different projections.}
	\label{fig:res}
\end{figure*}

\section{Introduction}
\label{sec:intro}

\quad Over the past few years, with the development of Virtual Reality (VR) technologies, VR has been widely used in many fields such as gaming, social networking, education, and medical care. In particular, omnidirectional video, also known as 360-degree video, is an elemental form of VR. In order to provide users with better immersive experience, higher resolution omnidirectional videos are strongly demanded, resulting in huge storage and computational cost compared with ordinary videos. Therefore, omnidirectional SR is crucial for VR technologies.

Unlike ordinary video, omnidirectional video cannot be encoded into 2D rectangular flat video directly due to its spherical characteristics \cite{Omnidirectional}. Therefore, before super-resolving the omnidirectional video, it is necessary to project the 3D sphere to a 2D plane. Various projection methods have been proposed such as Equi-Rectangular Projection (ERP), Ico-Sahedral Projection (ISP) \cite{ISP}, Octa-Hedron Projection (OHP), Cube-Map Projection (CMP), Truncated Square Pyramid Projection (TSP), Segmented Sphere Projection (SSP) and Equi-Angular cube map projection (EAC). 

Among these projection methods, ERP is the most intuitive projection, since the complete linear transformation formula makes it easy to operate in VR display. Therefore, existing methods on omnidirectional SR use ERP to project the omnidirectional video. \cite{deng2021lau} proposed a latitude adaptive upscaling network, which allows pixels at different latitudes to adopt distinct upscaling factors. They used a deep reinforcement learning scheme to automatically select optimal upscaling factors for different latitude bands. \cite{yoon2022spheresr} introduced a novel framework to predict the RGB values at given spherical coordinates for omnidirectional SR. They extracted the feature based on ISP and used a spherical local implicit image function to predict the RGB values. \cite{yoon2022spheresr} achieved better results compared with \cite{deng2021lau}. Although \cite{yoon2022spheresr,deng2021lau} are specially designed for omnidirectional SR, their SR networks are all constructed based on 2D convolution module, which is more suitable for the regular grid.

Among all the projection methods, ERP has severe projection distortion near poles, since the uniform sampling leads to higher sampling density at the poles compared with the equator. CMP is an uniform projection method compared with ERP, while it still contains certain distortion. EAC improves traditional CMP by adjusting the position of the up sampled pixels on the cube according to the spherical coordinates. As can be seen, various projections result in different distortions. Therefore, the characteristics of projection methods will have an impact on the performance of omnidirectional SR. 

To validate the impact of various projection methods, we compare the performances based on different SR architectures including EDSR \cite{edsr}, RCAN \cite{rcan} and SwinIR \cite{swinir} across different scaling factors (x2, x3, x4) and datasets. Experimental results show that Equi-Angular cube map projection (EAC), which has minimal distortion, achieves the best result in terms of WS-PSNR compared with other projections. We hope our comprehensive comparison can inspire future works on omnidirectional SR. 

Our contributions can be concluded as follows:
\begin{itemize}
	\item We analyze the distortion characteristics of various projection methods including ERP, ISP, OHP, CMP, TSP, SSP and EAC.
	\item We provide a comprehensive comparison of projections on omnidirectional SR, showing that minimal distortion projection leads to best SR performance.
	\item We conduct extensive experiments based on different SR architectures across various scaling factors, metrics and datasets.
\end{itemize}


\section{Proposed Method}
\label{PM}

\quad In this section, we propose our method to handle the omnidirectional super-resolution. The input to our method is the ERP low resolution frame, which is denoted as $I_{LR}^{erp}$. Our method uses different projections to project the omnidirectional videos including Equi-Rectangular Projection (ERP), Ico-Sahedral Projection (ISP) \cite{ISP}, Octa-Hedron Projection (OHP), Cube-Map Projection (CMP), Truncated Square Pyramid Projection (TSP), Segmented Sphere Projection (SSP) and Equi-Angular cube map projection (EAC). The illustration of different projections is shown in Fig. \ref{fig:res}. We denote the projection conversion as $T$. For example, we use ${T^{erp -  > eac}}$ to denote the projection conversion from ERP to EAC. We denote the SR networks trained on various projections as $N$. For example, the SR network trained on EAC can be represented as ${N^{eac}}$. Therefore, taking EAC as one example, the process of our method can be represented as follows.

\begin{table}[]
	\renewcommand\arraystretch{1.5}
	\caption{Quantitative results of different models.}
	\resizebox{\linewidth}{!}{
		\begin{tabular}{|l|l|l|l|l|l|l|l|l|}
			\hline
			&         & ERP    & ISP   & OHP    & TSP   & SSP    & CMP   & EAC    \\ \hline
			\multirow{3}{*}{EDSR}    & PSNR X2 & 29.098 & 24.73 & 25.987 & 22.93 & 24.597 & 31.36 & 34.18  \\ \cline{2-9} 
			& PSNR X3 & 23.91  & 20.44 & 21.78  & 18.91 & 20.90  & 24.07 & 26.99 \\ \cline{2-9} 
			& PSNR X4 & 21.34  & 19.41 & 20.558 & 17.78 & 19.94  & 20.59 & 22.72  \\ \hline
			\multirow{3}{*}{RCAN}    & PSNR X2 & 28.45  & 24.46 & 25.78  & 22.77 & 24.35  & 31.34 & 33.76  \\ \cline{2-9} 
			& PSNR X3 & 23.61  & 20.12 & 21.44  & 18.73 & 20.61  & 23.54 & 26.42  \\ \cline{2-9} 
			& PSNR X4 & 20.86  & 18.91 & 20.06  & 17.58 & 19.47  & 19.72 & 21.83  \\ \hline
			\multirow{3}{*}{SWIN IR} & PSNR X2 & 28.40  & 24.35 & 25.71  & 22.60 & 24.28  & 31.19 & 33.79  \\ \cline{2-9} 
			& PSNR X3 & 23..91 & 20.31 & 21..62 & 18.79 & 20.77  & 23.87 & 26.75  \\ \cline{2-9} 
			& PSNR X4 & 21.37  & 19.20 & 20..35 & 17.76 & 19.72  & 20.16 & 22.43  \\ \hline
		\end{tabular}
	}
	\label{tab:tab_1}
\end{table}

\begin{table}[htbp]
	\centering
	\renewcommand\arraystretch{1.5}{\centering}
	\scriptsize
	\caption{Quantitative results of WS-PSNR}
	\begin{tabular}{|c|c|c|c|}
		\hline
		~ & WS-PSNR X2 & WS-PSNR X3 & WS-PSNR X4 \\ \hline
		ERP & 28.925 & 24.306 & 21.976 \\ \hline
		ISP & 24.119 & 19.224 & 18,778 \\ \hline
		OHP & 24.059 & 19.11 & 18.805 \\ \hline
		SSP & 22.972 & 19.262 & 18.328 \\ \hline
		TSP & 21.766 & 18.847 & 18.199 \\ \hline
		CMP & 31.977 & 25.978 & 22.46 \\ \hline
		EAC & \textbf{33.524} & \textbf{28.653} & \textbf{24.454} \\ \hline
	\end{tabular}
	\label{tab:tab_3}
        \vspace{-1.2em}
\end{table}
\begin{equation}
\label{eq:1}
I_{SR}^{erp} = {T^{eac -  > erp}}(N({T^{erp -  > eac}}(I_{LR}^{erp})))
\end{equation}

As for the individual SR network on each projection, we train them separately as:

\begin{equation}
\label{eq:2}
I_{SR}^{eac} = N(I_{LR}^{eac};{\theta _N})
\end{equation}

where ${\theta _N}$ is the weights of EAC SR network. We use three prevailing SR architectures including EDSR \cite{edsr}, RCAN \cite{rcan} and SwinIR \cite{swinir} across different scaling factors (x2, x3, x4). We use the L1 loss between SR and HR to train the network:

\begin{equation}
\label{eq:3}
loss = {\left\| {I_{SR}^{eac} - I_{HR}^{eac}} \right\|_1}
\end{equation}

Other networks of different projections are trained similarly. The whole pipeline of our method is illustrated by Fig. \ref{fig:pipeline}.

\section{EXPERIMENTS}
\label{sec:majhead}

\begin{figure*}[!h]
	\centering
	\includegraphics[width=\textwidth,height=0.34\textwidth]{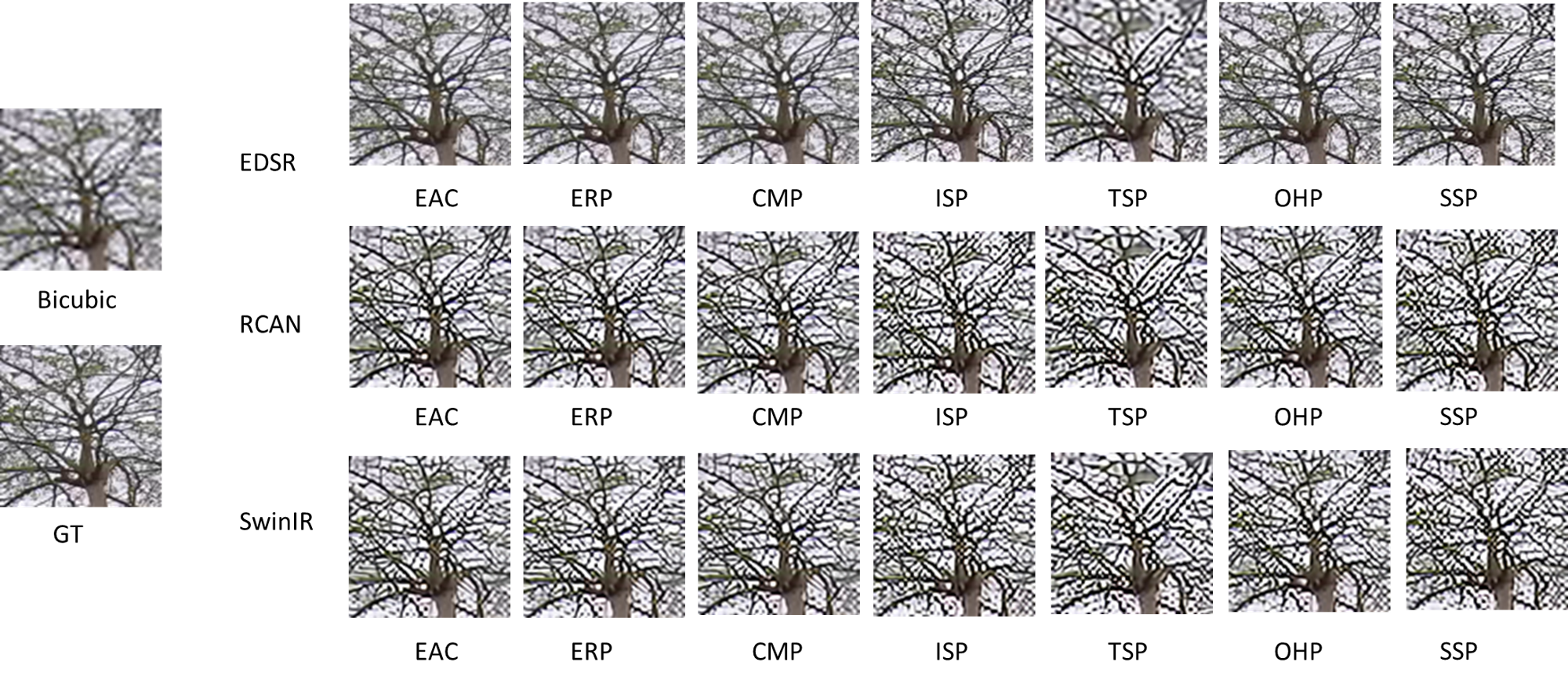}
	\includegraphics[width=\textwidth,height=0.34\textwidth]{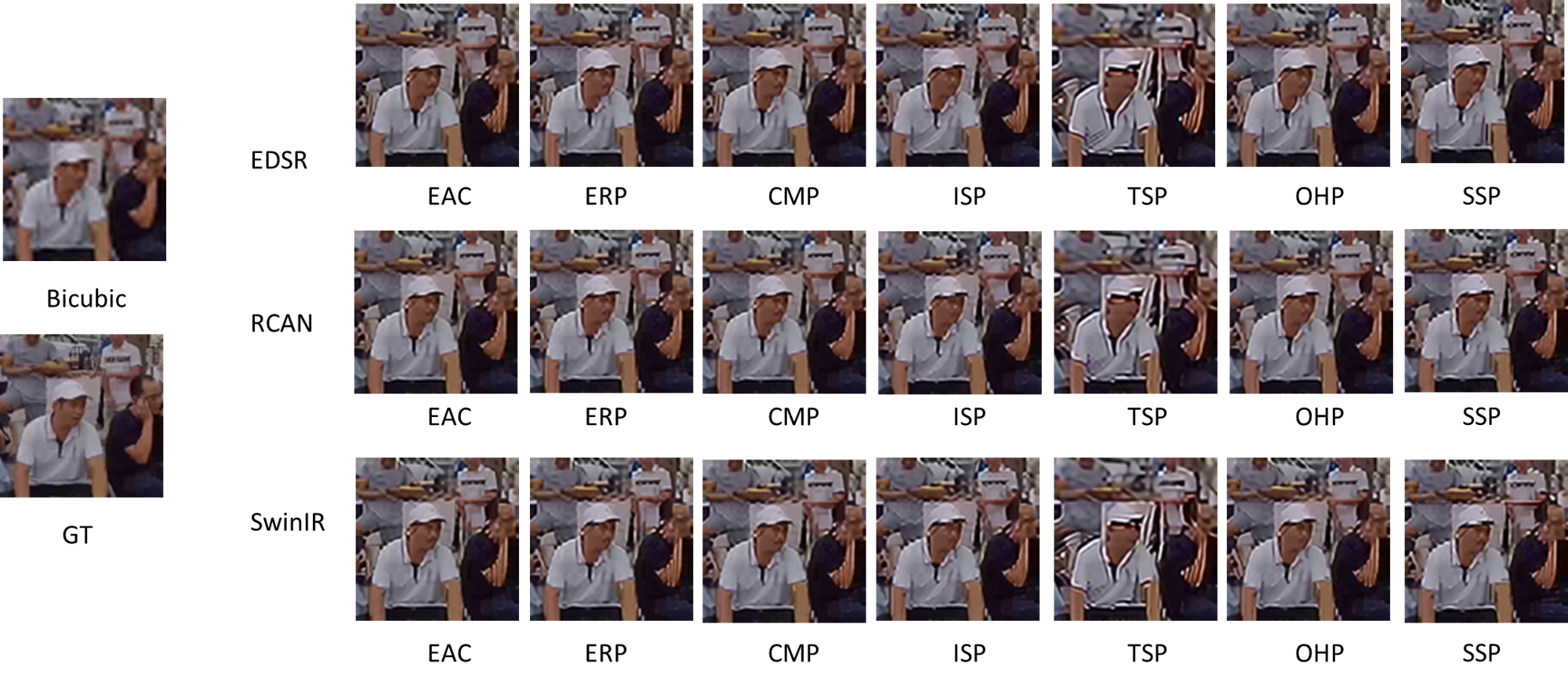}
	\caption{Qualitative Results.}
	\label{fig:qual_res}
\end{figure*}

\quad In this section, we introduce our experiments in detail, including experimental settings, evaluation indicators, experiment results and analysis.
\subsection{Experimental Settings}
\label{sec:exp_setting}

\quad We implement the experiments with PyTorch framework and use EDSR \cite{edsr}, RCAN \cite{rcan} and SwinIR \cite{swinir} as the baseline architectures. We use 360 Tools and FFmpeg  \cite{ffmpeg} for conversion between different projection formats. Besides, our experiments follow the Common Test Conditions (CTC) \cite{un}. As for the testing dataset, we test the projections on our own dataset and Fig. \ref{fig:res} is an example of our testing dataset. Firstly, we evaluate the PSNR of each projection based on the above three baselines. We also evaluate several widely-used objective evaluation indicators \cite{image_evaluation} based on EDSR, including Weighted to Spherically uniform PSNR (WS-PSNR) \cite{un} \cite{unknown}, Learned Perceptual Image Patch Similarity (LPIPS) \cite{lpips} metric and Structural Similarity (SSIM) \cite{ssim}.

\begin{table}[htbp]
	\renewcommand\arraystretch{1}
	\centering
        \vspace{-1.2em}
	\caption{Quantitative results of different metrics (on EDSR)}
	\resizebox{\linewidth}{!}{
		\begin{tabular}{|l|l|l|l|l|l|l|l|}
			\hline
			~ & ERP & ISP & OHP  & TSP  & SSP & CMP  & EAC \\ \hline
			PSNR X2 & 29.098 & 24.73 & 25.987 & 22.93 & 24.597 & 31.36 &\textbf{34.18} \\ \hline
			SSIM X2 & 0.948 & 0.912 & 0.934 & 0.834 & 0.916 & 0.957 & \textbf{0.976} \\ \hline
			LPIPS X2 & 0.0904 & 0.114 & 0.0939 & 0.1962 & 0.1068 & 0.0703 & \textbf{0.0556} \\ \hline
			~ & ~ & ~ & ~ & ~ & ~ & ~ & ~ \\ \hline
			PSNR X3 & 23.91 & 20.44 & 21.78 & 18.91 & 20.9 & 24.07 &\textbf{26.989} \\ \hline
			SSIM X3 & 0.884& 0.813 & 0.855 & 0.677 & 0.836 & 0.831 & \textbf{0.917} \\ \hline
			LPIPS X3 & 0.150 & 0.193 & 0.172 & 0.346 & 0.168 & 0.186 &\textbf{0.135}\\ \hline
			~ & ~ & ~ & ~ & ~ & ~ & ~ & ~ \\ \hline
			PSNR X4 & 21.34 & 19.41 & 20.558 & 17.78 & 19.94 & 20.59 & \textbf{22.72} \\ \hline
			SSIM X4 & 0.827 & 0.769 & 0.815 & 0.62 & 0.803 & 0.705 & \textbf{0.844} \\ \hline
			LPIPS X4 & 0.209 & 0.245 & 0.2098 & 0.424 & 0.209 & 0.272 & \textbf{0.199} \\ \hline
		\end{tabular}
	}
\label{tab:tab_2}
\vspace{-1.2em}
\end{table}
\subsection{Quantitative Results of Different Models}
\label{sec:quanti_models}

\quad In order to evaluate the performances of different projections, we design experiments to compare ERP, CMP and five other projection methods. Tab. \ref{tab:tab_1} shows the performances of different projection formats on various super-resolution networks. Tab. \ref{tab:tab_2} presents PSNR, SSIM and LPIPS of different projections on EDSR. As can be seen, Equi-Angular Cube map (EAC) projection has superior quality compared with other projections across scaling factors (x2, x3, x4). We investigate and analyze the reasons as below.

ERP is the most intuitive projection method since it can be easily transformed to display in the VR devices. However, ERP has severe projection distortion near poles. Compared with ERP, CMP maintains better image continuity since adjacent regions in CMP are also adjacent on the cube map. Therefore, CMP can achieve better performance compared with ERP. EAC further improves CMP by sampling more pixels at the edge area than the central area. This is because the edge area of cube map is sparse while the center area of cube map is sparse. EAC improves this uneven distribution by adjusting the positions of sampling pixels on the cube map corresponding to the spherical coordinates.

Apart from ERP and cube map (CMP and EAC), TSP can be well adapted to the aspect ratios of videos and is easy to encode since all the frame is effective pixels. Although all sides of TSP are continuous, severe image distortion exists in each side. SSP is a simple method and the projection effect is relatively smooth. However, the uniformity of SSP is poor and there is a gray default area, which is a waste of storage space. ISP and OHP use regular octahedron and icosahedron separately instead of cube map as the projection models to improve the sampling uniformity. However, their per-face boundary artifacts and missing pixels during compression degrade the SR performance to some extent.

\subsection{Quantitative Results of Projection Conversion}
\label{sec:proj_conv}

\quad In order to perform a more thorough comparison between different projections, we conduct the quantitative comparison of projection conversion. To be more specific, we first convert the original HR ERP videos to LR projections such as EAC, CMP, SSP and so on. Then we perform super-resolution at different scales with EDSR and back-project the SR results (EAC, CMP, SSP, etc.) to generate SR ERP outputs. Finally, we evaluate the WS-PSNR between SR ERP outputs and HR ERP ground-truth. WS-PSNR is a widely-used metric in omnidirectional SR. We perform WS-PSNR evaluation on the three channels of YUV and report the results on Y channel. The WS-PSNR results of different projections are shown in Tab. \ref{tab:tab_3}. It can be seen from the results that EAC has higher quality than other advanced projection methods across different scaling factors. cube map projections like CMP and EAC achieve better performance than widely-used ERP \cite{deng2021lau,yoon2022spheresr}. In particular, EAC achieves a gain of 4.60 dB on the X2 scale, 4.347 dB on the X3 scale, and 2.478 dB on the X4 scale compared with ERP. The CMP also achieves a gain of 3.052 dB and 1.672 dB compared with ERP on the X2 and X3 scales. Other projection methods such as ISP, OHP, etc. have caused a certain degree of decline. 

\subsection{Qualitative Results}
\label{sec:qual_res}

\quad Fig. \ref{fig:qual_res} shows the qualitative comparison of different projections. Similar to projection conversion comparison, we first convert the ERP input to different projections and super-resolve them. Then we convert the SR results to SR ERP output. Finally, we show the qualitative results of different projections for a fair comparison. As we can see, EAC achieves the best results compared with other projection methods since it contains minimal distortion.

\section{CONCLUSION AND DISCUSSION}
\label{sec:conc}

\quad In this paper, the performances of various omnidirectional projections are compared, among which EAC achieves the best performance since it contains the minimal distortion. We hope our comprehensive comparison can inspire future works on omnidirectional super-resolution. As for future work, we suggest two promising directions: 1) Apart from EAC, there are some other better projection methods that are more suitable for omnidirectional super-resolution. For example, Hybrid Equiangular cube map (HEC), which maintains good pixel continuity and uniformity during the projection process, will also help improve the SR performance. 2) Applying our work to reduce the bandwidth of VR video transmission is also a promising and important direction.

\vfill\pagebreak

\bibliographystyle{IEEEbib}
\bibliography{strings,refs}

\end{document}